\documentclass[11pt]{article}

\usepackage[preprint]{acl}

\usepackage{times}
\usepackage{latexsym}
\usepackage{amsmath}
\usepackage{booktabs}
\usepackage{stfloats}
\usepackage{placeins}
\usepackage{booktabs}
\usepackage{multirow}
\usepackage{float}
\usepackage[dvipsnames]{xcolor}
\usepackage{textcomp}
\usepackage{amssymb}

\usepackage[most]{tcolorbox}
\tcbuselibrary{skins, breakable, raster}
\usepackage{graphicx}
\usepackage{colortbl}
\usepackage{array}
\usepackage{tabularx}
\usepackage{cuted}
 
\definecolor{claimgold}{RGB}{255,248,220}
\definecolor{claimframe}{RGB}{180,140,60}
\definecolor{chartbg}{RGB}{235,242,255}
\definecolor{chartframe}{RGB}{80,110,180}
\definecolor{tablebg}{RGB}{235,250,240}
\definecolor{tableframe}{RGB}{60,150,90}
\definecolor{correct}{RGB}{34,139,34}
\definecolor{incorrect}{RGB}{178,34,34}
\definecolor{rowrule}{RGB}{180,180,180}
 
\newcommand{\correctbadge}[1]{%
  \colorbox{correct!15}{\textcolor{correct}{\textbf{\checkmark}~#1}}}
\newcommand{\incorrectbadge}[1]{%
  \colorbox{incorrect!12}{\textcolor{incorrect}{\textbf{\texttimes}~#1}}}

\usepackage[T1]{fontenc}

\usepackage[utf8]{inputenc}

\usepackage{microtype}

\usepackage{inconsolata}

\usepackage{graphicx}

\title{Encoded but Not Routed: Explaining the Table-Chart Gap \\ in Scientific Claim Verification}

\author{
  \textbf{Sunisth Kumar}$^{1,2}$ \quad
  \textbf{Xanh Ho}$^{2}$ \quad
  \textbf{Tim Schopf}$^{3}$ \\
  \textbf{Andre Greiner-Petter}$^{3,4}$ \quad
  \textbf{Florian Boudin}$^{5}$ \quad
  \textbf{Akiko Aizawa}$^{1,2,3}$ \\[4pt]
  $^{1}$The University of Tokyo \quad
  $^{2}$NII LLMC \quad
  $^{3}$National Institute of Informatics \\
  $^{4}$University of Göttingen \quad
  $^{5}$Inria, LS2N, Nantes Universit\'{e} \\[4pt]
  \small
  \texttt{sunisth@g.ecc.u-tokyo.ac.jp} \quad
  \texttt{\{xanh,aizawa\}@nii.ac.jp} \quad
  \texttt{tim.schopf@t-online.de} \\
  \small
  \texttt{greinerpetter@gipplab.org} \quad
  \texttt{florian.boudin@univ-nantes.fr}
}

\begin{document}
\maketitle
\begin{abstract}
Multimodal LLMs are increasingly used to assist scientific peer review, where a core requirement is verifying whether claims in a paper are supported by its evidence. Prior work has shown that models perform substantially better at this task when the evidence is a table than when it is a chart of the same underlying data. This raises the question of whether models fail to extract information from charts, or do they extract it but fail to use it when forming their prediction? We study this question through layer-wise linear probing and attention analysis on three open-weight VLMs over table and chart evidence, representing the same underlying data. We find consistent evidence for the latter. Chart information is encoded in the models' intermediate representations but does not reach the prediction position, a gap that is absent for tables and holds across all conditions tested. Attention analysis further reveals that this disconnect takes two architecturally distinct forms across model families. These findings point toward reframing the table-chart gap as a failure of how encoded visual information is used at prediction time, rather than a failure of encoding itself. \footnote{\url{https://github.com/ksunisth/encoded-not-routed}}
\end{abstract}

\section{Introduction}

\begin{figure}[t]
    \centering
    \includegraphics[width=0.85\columnwidth]{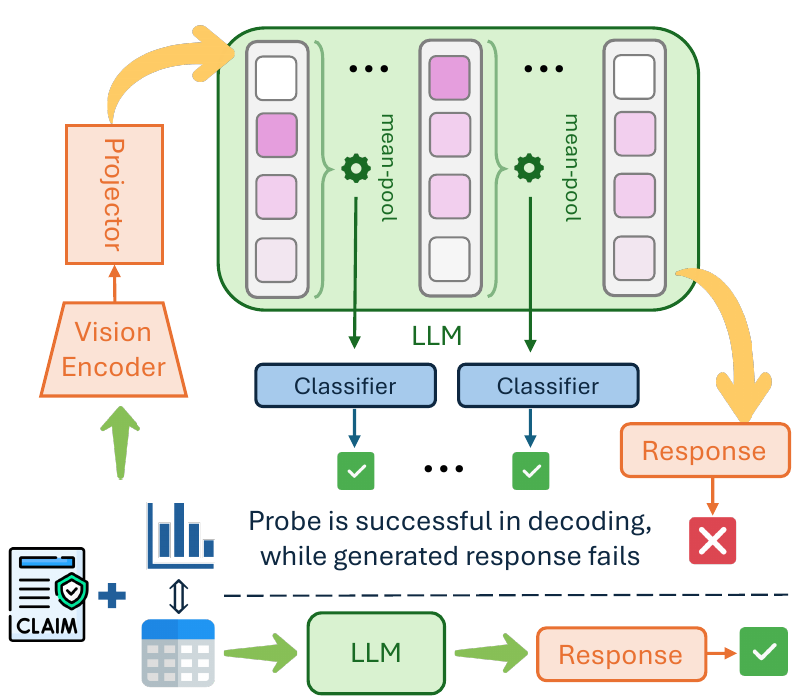}
    \caption{Overview of our approach. We apply linear probing to compare how chart and table evidence is encoded and passes to the model response.}
    \label{fig:overview}
\end{figure}

Multimodal large language models are increasingly used for scientific claim verification, a task that requires determining whether a given claim is supported or refuted by evidence from research papers \cite{lu-etal-2023-scitab, wang-etal-2025-sciver, ho2026sciclaimevalcrossmodalclaimverification}. In practice, the same experimental finding may be presented as a table in one case and as a bar or line chart in another. A robust verifier should therefore make consistent decisions across evidence formats when the underlying data is unchanged.

Recent work shows that current multimodal LLMs perform substantially better on tables than on equivalent charts \cite{Ho_Wu_Kumar_Boudin_Takasu_Aizawa_2026}. Since the claim and the underlying data are held constant across formats, this reveals a fundamental brittleness in how models process semantically equivalent information in different modalities. Understanding why this gap exists matters because it determines whether models fail to extract visual information or fail to use information they have already extracted.

What explains this asymmetry? One possibility is that models fail to perceive chart content, such that visual information does not reach model representations in a usable form. Another is that chart content is encoded in model representations but is not effectively routed to the prediction stage. These two explanations call for different responses. Perceptual failures point toward better vision encoders or chart-specific pre-training \cite{masry-etal-2022-chartqa, liu-etal-2025-perception}, while utilization failures call for interventions on how encoded representations are used at prediction time \cite{Zhang_2025_CVPR, kawasaki2026responsesfallshortunderstanding}. Distinguishing between them requires looking inside the model.

We take a diagnostic approach to this question, studying three open-weight multimodal LLMs from two model families on a content-controlled SciTabAlign+ benchmark \cite{Ho_Wu_Kumar_Boudin_Takasu_Aizawa_2026} in which each claim is paired with semantically equivalent table and chart evidence. We analyze internal representations through linear probing \cite{belinkov-2022-probing} to understand whether chart failures arise from missing internal signals or from failures to use that signal during prediction (see Figure~\ref{fig:overview}).

Our results reveal that chart-relevant signal is recoverable from intermediate representations yet is not effectively routed to final verification decisions, indicating a routing failure rather than a perceptual one.

\section{Related Work}
\textbf{Scientific Claim Verification.} 
Claim verification has been studied from textual evidence \cite{thorne-etal-2018-fever}, and extended to the scientific domain through SciFact \cite{wadden-etal-2020-fact}, SciTab \cite{lu-etal-2023-scitab}, and SciTabAlign \cite{ho-etal-2025-table}. Recent works have broadened the task to multimodal settings spanning tables and figures \cite{lal-etal-2025-musciclaims, wang-etal-2025-sciver, ho2026sciclaimevalcrossmodalclaimverification}, where multimodal LLMs have been shown to perform substantially better on tables than on equivalent charts \cite{Ho_Wu_Kumar_Boudin_Takasu_Aizawa_2026}. Building on this finding, we investigate why that gap arises.

\textbf{Chart Understanding and Interpretability in Multimodal LLMs.}
ChartQA \cite{masry-etal-2022-chartqa} and CharXiv \cite{NEURIPS2024_cdf6f8e9} benchmark VLM performance on chart-specific tasks, both revealing persistent gaps between VLM and human performance. \cite{11261397} further apply activation maps to visualize which chart regions models attend to. Unlike these studies, we hold the underlying data constant across formats, isolating format sensitivity from content variation. 
Recent work shows VLM failures may arise even when relevant visual information is present in representations \cite{kawasaki2026responsesfallshortunderstanding}, and that models may attend to relevant visual evidence while still answering incorrectly \cite{liu2025seeingbelievingprobingdisconnect, Esmaeilkhani_2026_WACV}. We study this representation-response gap in a format-controlled claim verification setting, where the same underlying data appears as table or chart evidence.

\section{Experimental Setup}
\paragraph{Task and Data.}
We study binary scientific claim verification on SciTabAlign+. The dataset pairs 162 unique scientific claims with a structured table and four chart variants (basic bar chart, symbol bar chart, line chart, and swapped chart), all drawn from the same underlying data. In the dataset, tables are provided as JSON and charts as JPG images. Each claim is labeled as \textit{Supported} or \textit{Refuted}, resulting in 372 table-based and 648 chart-based instances. Holding evidence content constant across formats allows us to attribute behavioral differences to format rather than to content.

\paragraph{Models.}
We study three open-weight multimodal LLMs spanning two families. We use Qwen2.5-VL-7B-Instruct and Qwen2.5-VL-32B-Instruct from the Qwen family \cite{bai2025qwen25vltechnicalreport}, and InternVL3-8B from the InternVL family \cite{zhu2025internvl3exploringadvancedtraining}. 

\paragraph{Linear Probing.}
We train a linear probe at every layer $l$ of each model using leave-one-out cross-validation over the 162 claims, minimizing binary cross-entropy with Adam~\cite{kingma2015adam} (learning rate 0.01, 300 steps). Each probe maps a hidden state $\mathbf{h}^{(l)}$ to a label prediction: 
\begin{equation}
    p\!\left(y = 1 \mid \mathbf{h}^{(l)}\right) =
    \sigma\!\left(\mathbf{w}^\top \mathbf{h}^{(l)} + b\right)
    \label{eq:probe}
\end{equation} 
Within each fold, we determine the probe's direction from a stratified 20\% inner split of the training data. We evaluate each probe under two settings. In the \textit{last-token} setting, $\mathbf{h}^{(l)}$ is the hidden state at the final input token position, from which the model generates its prediction. In the \textit{mean-pool} setting, $\mathbf{h}^{(l)}$ is the mean hidden state averaged uniformly over all input token positions, which captures task-relevant information distributed across layers rather than concentrated at the final position \cite{layer_by_layer}.
As a control for claim-text leakage, we train identical probes on claim text alone, passing each claim through the model without evidence input. See Appendix~\ref{app:claim-only}.

\paragraph{Attention Analysis.} At each layer $l$, we compute the fraction of last-token attention directed to image token positions, $a^{(l)}_\text{img}$, normalized by the proportional baseline:
\begin{equation}
    \hat{a}^{(l)} = \frac{a^{(l)}_\text{img}}{|T_\text{img}| \,/\, |T|}
    \label{eq:attention}
\end{equation}
where $|T_\text{img}|$ is the number of image tokens and $|T|$ is the total token count. We compute $a^{(l)}_\text{img}$ by averaging attention uniformly across heads and summing across all image-token positions, which measures total probability mass directed to the image region. A value of $\hat{a}^{(l)}$ = $1.0$ means the model attends to image tokens in proportion to their count in the input. Values below $1.0$ indicate less attention to image content than this baseline.

\paragraph{Chain-of-Thought Ablation.}
We test whether explicit chart verbalization before prediction can improve performance accuracy by prompting each model to describe the chart before predicting the label (see Appendix~\ref{app:cot}).

\paragraph{Evaluation.}
We evaluate probing using area under the receiver operating characteristic curve (AUROC). Our interest is in whether task-relevant information is decodable from representations, and AUROC captures this without relying on a fixed classification threshold. We also report macro-F1 and accuracy for direct comparison with model output accuracy.

\section{Results}

\begin{figure}[t]
    \centering
    \includegraphics[width=\columnwidth]{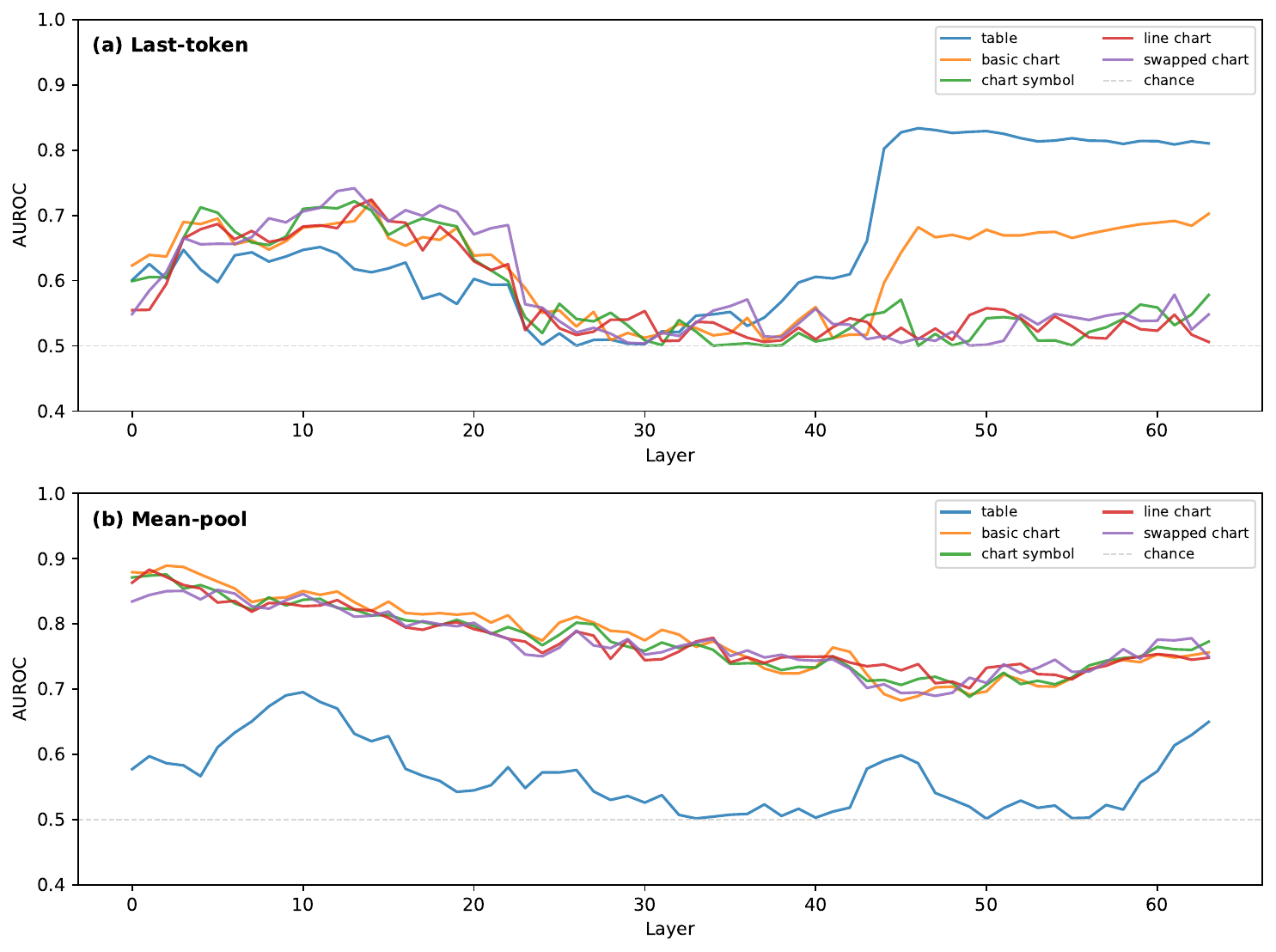}
    \caption{Probe AUROC by layer for Qwen2.5-VL-32B on SciTabAlign+. \textbf{(a)} Last-token: table rises sharply in late layers, chart variants peak early and fall toward chance in late layers. \textbf{(b)} Mean-pool: chart signal remains decodable across all layers, but does not reach prediction position. Results for all models are shown in Appendix~\ref{app:probing-auroc}.}
    \label{fig:probing-curves}
\end{figure}

\paragraph{Chart signal is recoverable from intermediate representations.}
Figure~\ref{fig:probing-curves} shows that mean-pooled probes substantially outperform last-token probes on chart variants across all models. For Qwen2.5-VL-32B, mean-pool AUROC on basic chart reaches 88.9\% compared to 72.2\% at the last token, indicating that chart-relevant signal is distributed across hidden states but has not concentrated at the prediction position. Notably, mean-pool AUROC is higher for chart variants than for table evidence across all models (84--89\% vs. 65--70\%), the reverse of the last token pattern in Table~\ref{tab:probing}. This inversion suggests that chart information is more broadly distributed across hidden states than table information, but specifically fails to concentrate at the prediction position where table information already aggregates. Claim-only probes trained on claim text without evidence peak at 58--67\% AUROC across models (see Appendix~\ref{app:claim-only}), well below the mean-pool AUROC on chart variants, ruling out claim-text leakage and indicating that the probe captures information present in image tokens but absent from claim text alone. 
Table and chart also converge to high cosine similarity in deeper layers (see Appendix~\ref{app:convergence}), ruling out representational divergence as the source of the behavioral gap. 

\begin{figure*}[ht]
    \centering
    \includegraphics[width=\textwidth]{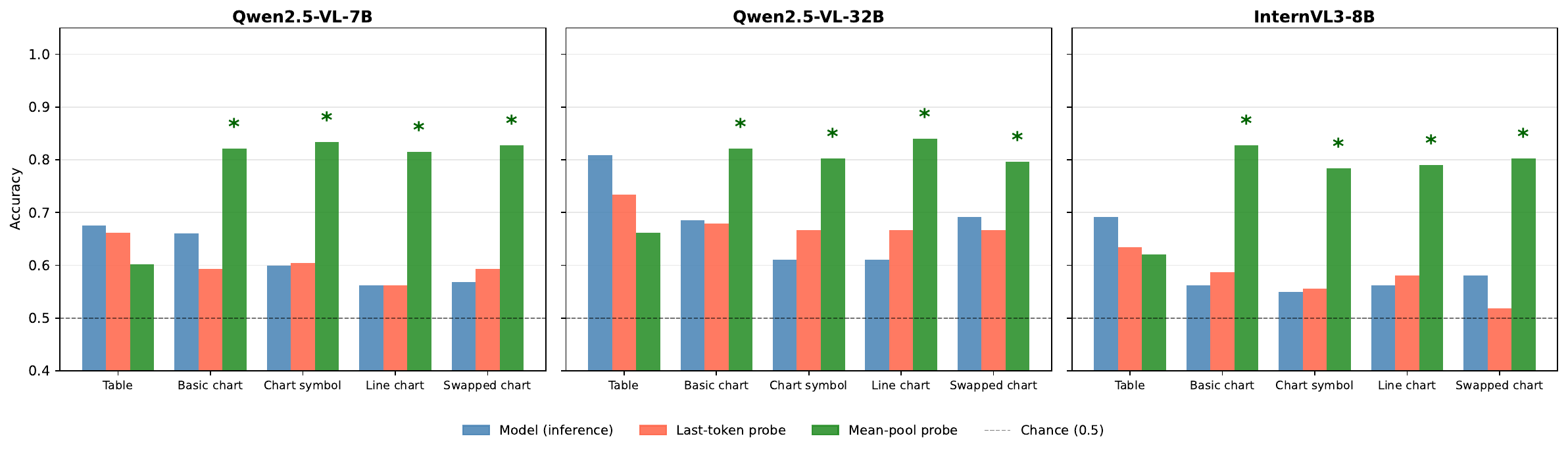}
    \caption{Probe accuracy vs. model inference accuracy on SciTabAlign+. The $*$ indicates chart variants where the mean-pool probe exceeds model accuracy.}
    \label{fig:probe-classifier}
\end{figure*}

\paragraph{Encoded signal does not reach the prediction position.}
Table~\ref{tab:probing} shows that last-token AUROC is overall consistently higher for table evidence than for chart evidence across all models and chart variants. 
As shown in Figure~\ref{fig:probe-classifier}, a linear classifier trained on mean-pooled representations exceeds model inference accuracy on all chart variants across all three models, a pattern absent for table evidence. The dissociation is significant (McNemar's test \cite{McNemar1947}, $p < 0.05$ in all conditions and $p < 0.01$ in all but one) and reverses on table evidence, confirming format specificity. For InternVL3-8B, the mean-pool probe achieves an accuracy of 82.7\% on basic chart while the model itself reaches 56.2\%. This dissociation, where a simple linear classifier outperforms the model using its own representations, suggests that chart-relevant information is present but is not effectively leveraged at prediction time. The failure is therefore not one of encoding but of how encoded information is used when forming a prediction.

\begin{table}[ht]
\centering
\resizebox{0.99\columnwidth}{!}{%
\begin{tabular}{lrrrrrr}
    \toprule
    & \multicolumn{2}{c}{\textbf{Qwen2.5-VL-7B}}
    & \multicolumn{2}{c}{\textbf{Qwen2.5-VL-32B}}
    & \multicolumn{2}{c}{\textbf{InternVL3-8B}} \\
    \cmidrule(lr){2-3}\cmidrule(lr){4-5}\cmidrule(lr){6-7}
    \textbf{Format} & AUROC & F1 & AUROC & F1 & AUROC & F1 \\
    \midrule
    Table          & \textbf{72.7} & \textbf{66.1}
                   & \textbf{83.4} & \textbf{75.2}
                   & \textbf{72.2} & \textbf{64.8} \\
    \midrule
    Basic chart    & 61.9 & 59.0 & 72.2 & 67.6 & 63.3 & 58.5 \\
    Chart symbol   & 62.0 & 60.3 & 72.2 & 70.2 & 58.8 & 55.5 \\
    Line chart     & 57.2 & 56.1 & 72.4 & 67.7 & 61.8 & 62.8 \\
    Swapped chart  & 60.8 & 59.1 & 74.2 & 67.4 & 56.8 & 53.7 \\
    \bottomrule
\end{tabular}%
}
\caption{Best last-token probe AUROC and macro-F1 across all layers on SciTabAlign+.}
\label{tab:probing}
\end{table}

\begin{figure}[ht]
    \centering
    \includegraphics[width=\columnwidth]{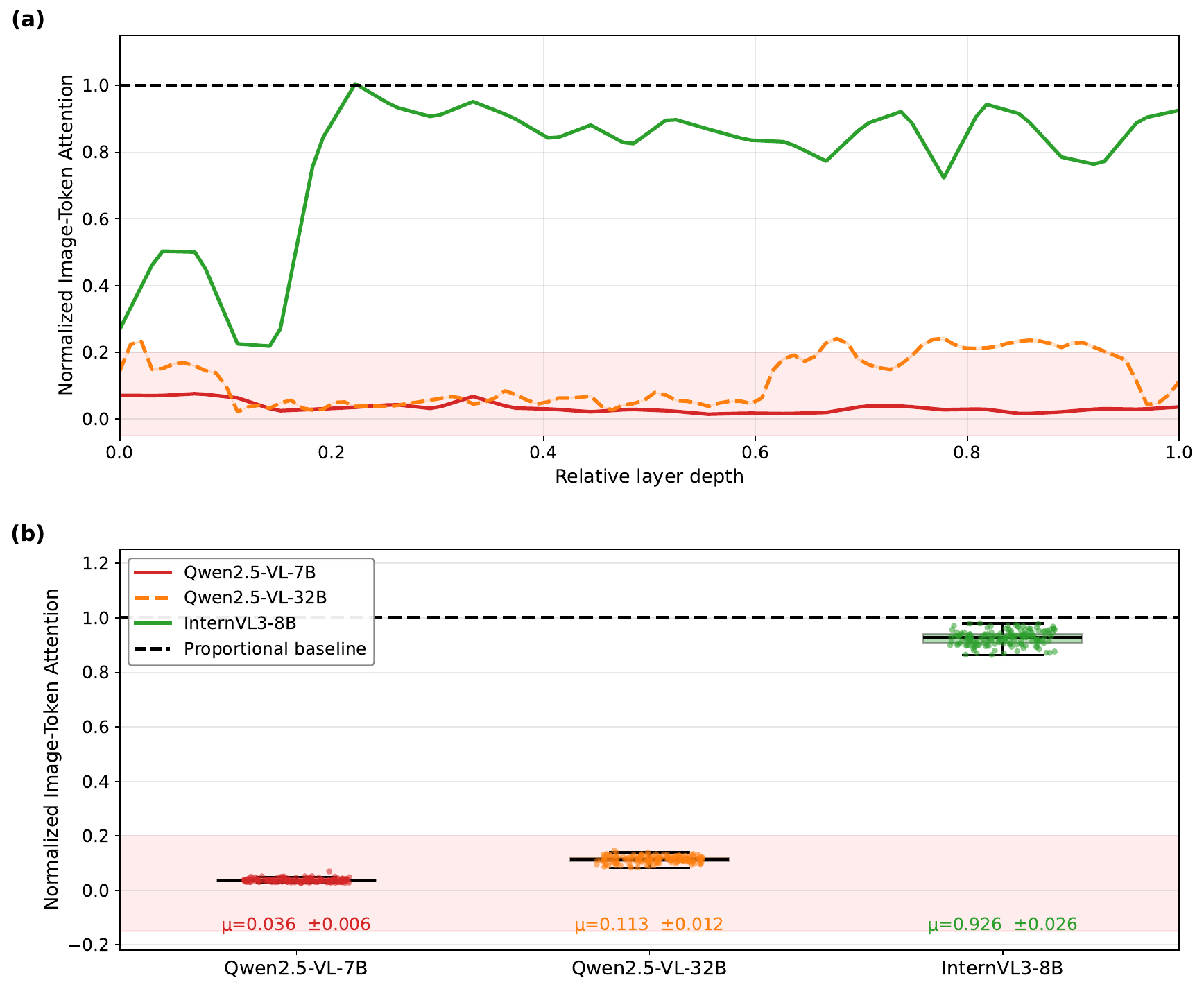}
    \caption{Image-token attention relative to the proportional baseline across layers on SciTabAlign+ (\textit{basic chart} variant). The shaded region indicates attention below 20\% of the proportional baseline. \textbf{(a)} Attention trajectories across layers. \textbf{(b)} Distribution of final-layer relative attention per claim.}
    \label{fig:attention}
\end{figure}

\paragraph{Failure patterns differ between Qwen and InternVL.}
Figure~\ref{fig:attention} shows that chart underutilization takes different forms across the two model families. Qwen-family models attend to image tokens only 4--11\% of the proportional baseline at the final layer, suggesting an \textit{attention routing failure} in which chart evidence is largely bypassed when predicting the claim label. InternVL3-8B, by contrast, maintains near-proportional image-token attention throughout most of the layers (93\% of the proportional baseline), yet still fails to outperform its own mean-pool probe on chart variants, consistent with a \textit{post-attention integration failure}, where the model attends to chart evidence but does not integrate it into the prediction. As shown in Appendix~\ref{app:internvl}, image-token attention distributes across positions rather than concentrating in summary tokens, confirming distributed visual routing.
As shown in Table~\ref{tab:cot-comparison}, chain-of-thought prompting further distinguishes the two families. CoT worsens macro-F1 for both Qwen models ($-13.5$ for 7B and $-3.3$ for 32B on basic chart), suggesting that forcing verbalization of chart content that the model has not attended to yields unreliable descriptions and hallucination. Qualitative analysis of model responses further corroborates the failure patterns (see Appendix~\ref{app:error-analysis}). InternVL3-8B benefits from CoT ($+4.6$ macro-F1), suggesting that model can access and describe chart content but does not draw on it spontaneously when predicting the claim label. The two failure modes are distinct in attention mechanism and response behavior, as observed in the opposite CoT effects across model families (Table~\ref{tab:cot-comparison}), indicating that the same intervention is unlikely to be equally effective for both.

\FloatBarrier
\section{Conclusion}
We investigated why multimodal LLMs perform substantially worse on chart evidence than on table evidence in scientific claim verification, despite the identical underlying data. Through layer-wise probing and attention analysis on three open-weight models, we find that chart-relevant signal is encoded and recoverable from intermediate representations but does not reach the prediction position, indicating a routing failure rather than a perceptual one. Attention analysis further reveals two distinct failure modes. Qwen-family models exhibit an \textit{attention routing failure}, largely bypassing chart evidence when predicting the claim label, while InternVL3-8B shows a \textit{post-attention integration failure}, attending to chart evidence but not integrating it into the prediction. Notably, the two model families with fundamentally different internal failures both show substantially lower accuracy on chart than on table evidence. Output evaluation alone cannot surface this distinction, underscoring that diagnosing format sensitivity requires examining internal computation, not only task performance.

\section{Limitations}
Our study has the following limitations: First, we use SciTabAlign+ dataset, which contains only 162 unique claims. While this is relatively small, the controlled design, where identical underlying data appears as tables and charts, enables us to isolate format effects that larger uncontrolled datasets cannot capture. The consistency of our findings across model families and chart variants provides confidence in the main result, despite the dataset size. The reported values are taken at the best layer per condition, selected on the test set, which can optimistically bias absolute values. The contrast between mean-pool and last-token decodability on chart evidence, however, holds at every layer.

Second, we analyze only open-weight multimodal LLMs because our analysis requires access to layer-wise hidden states and attention weights. This excludes proprietary models that are widely used in practice. We cannot determine whether the same internal routing failures appear in closed-source models.

Third, our analyses are diagnostic rather than causal. Linear probing and attention patterns provide correlational evidence of how information flows through models, but do not prove models causally use chart information during prediction. We demonstrate that chart information is recoverable from representations but is not reliably reflected in the model's final verification decision.

Finally, we focus on binary scientific claim verification task. It remains unclear whether similar format-specific failures appear in other tasks. Future work should test generalization beyond this setting.

Despite these limitations, our controlled setting makes the main finding informative. Even when underlying evidence data remains unchanged, chart-relevant information can be recoverable from model representations while still failing to support the model's final verification decision.

\section{Ethical Considerations}
This study uses only publicly available models and dataset, distributed under their respective licenses for research use, and involves no human subjects or new data collection. Our findings highlight that format-sensitive failures in multimodal LLMs may not be visible from model outputs alone. Researchers and practitioners deploying such models should treat model predictions with caution and use diagnostic analyses when evaluating systems for high-stakes scientific workflows. We used AI tools to assist with editing and coding in the preparation of this paper. All content has been thoroughly reviewed and verified by the human authors. 

\section{Acknowledgements}
This work was supported by JSPS KAKENHI Grant Number 24K03231, the Deutsche Forschungsgemeinschaft (DFG, German Research Foundation) – \href{https://gepris.dfg.de/gepris/projekt/554559555}{554559555}, and the German Academic Exchange Service (DAAD) - 57557629.

\bibliography{custom}

\appendix
\label{sec:appendix}

\section{Experimental Details}
\label{app:experimental-details}
All experiments were conducted on NVIDIA A100 (80GB) GPUs. Qwen2.5-VL-32B was distributed across 2 GPUs, while the 7B and 8B models each ran on a single GPU. Linear probes were implemented in PyTorch and trained from zero initialization on standardized features. Hidden states and attention weights were extracted using the Hugging Face Transformers library \cite{wolf-etal-2020-transformers}. Under 1000 label permutations, the probe accuracy stays at chance (0.500 ± 0.047, maximum 0.660), far below the observed accuracy ($p < 0.001$). An $\ell_2$-regularized scikit-learn \cite{scikit-learn} probe changes accuracy by at most 4.3 points and leaves all comparisons unchanged.

\section{Layer-wise Probing Analysis - All Models}
\label{app:probing-curves}

\subsection{Baseline Model Performance}
\label{app:baseline}

Table~\ref{tab:baseline-acc} shows baseline model performance. Models perform substantially better on tables than on charts, despite identical underlying data.

\begin{table}[H]
    \centering
    \resizebox{\columnwidth}{!}{%
    \begin{tabular}{lrrrrrr}
    \toprule
    & \multicolumn{2}{c}{\textbf{Qwen2.5-VL-7B}}
    & \multicolumn{2}{c}{\textbf{Qwen2.5-VL-32B}}
    & \multicolumn{2}{c}{\textbf{InternVL3-8B}} \\
    \cmidrule(lr){2-3}\cmidrule(lr){4-5}\cmidrule(lr){6-7}
    \textbf{Format} & Acc & F1 & Acc & F1 & Acc & F1 \\
    \midrule
    Table         & \textbf{67.5} & \textbf{70.5} & \textbf{80.9} & \textbf{83.8} & \textbf{69.1} & \textbf{70.4} \\
    \midrule
    Basic chart   & 66.0 & 65.9 & 68.5 & 68.2 & 56.2 & 55.2 \\
    Chart symbol  & 59.9 & 58.7 & 61.1 & 59.6 & 54.9 & 50.1 \\
    Line chart    & 56.2 & 55.5 & 61.1 & 60.0 & 56.2 & 53.0 \\
    Swapped chart & 56.8 & 56.7 & 69.1 & 68.8 & 58.0 & 56.9 \\
    \bottomrule
    \end{tabular}%
    }
    \caption{Model inference accuracy and macro-F1 (\%) on SciTabAlign+. Results use the baseline prompt tempate shown in Figure~\ref{app:baseline-prompt} .}
    \label{tab:baseline-acc}
\end{table}

\begin{figure}[H]
\noindent\fbox{%
\parbox{0.97\columnwidth}{%
\centering\textbf{Baseline Prompt Template}\\[0.5em]
\small
\begin{flushleft}
\textbf{System:} You are an expert in claim verification against scientific papers.\\[0.5em]
 
\textbf{User:} [image]\\
Use the provided image, predict the label for this claim: `\{claim\}'; the label can be Supported or Refuted. Think step by step before answering. Please format your final answer within brackets as follows: \texttt{<ans> YOUR ANSWER </ans>}
\end{flushleft}
}%
}
\caption{Baseline prompt template for claim verification. For the table format, [image] placeholder is replaced with [table] and the table text is prepended as
\texttt{Table Information: \{table\_text\}}. .}
\label{app:baseline-prompt}
\end{figure}

\subsection{AUROC Divergence}
\label{app:probing-auroc}
\begin{figure*}[t]
\centering
\includegraphics[width=\textwidth]{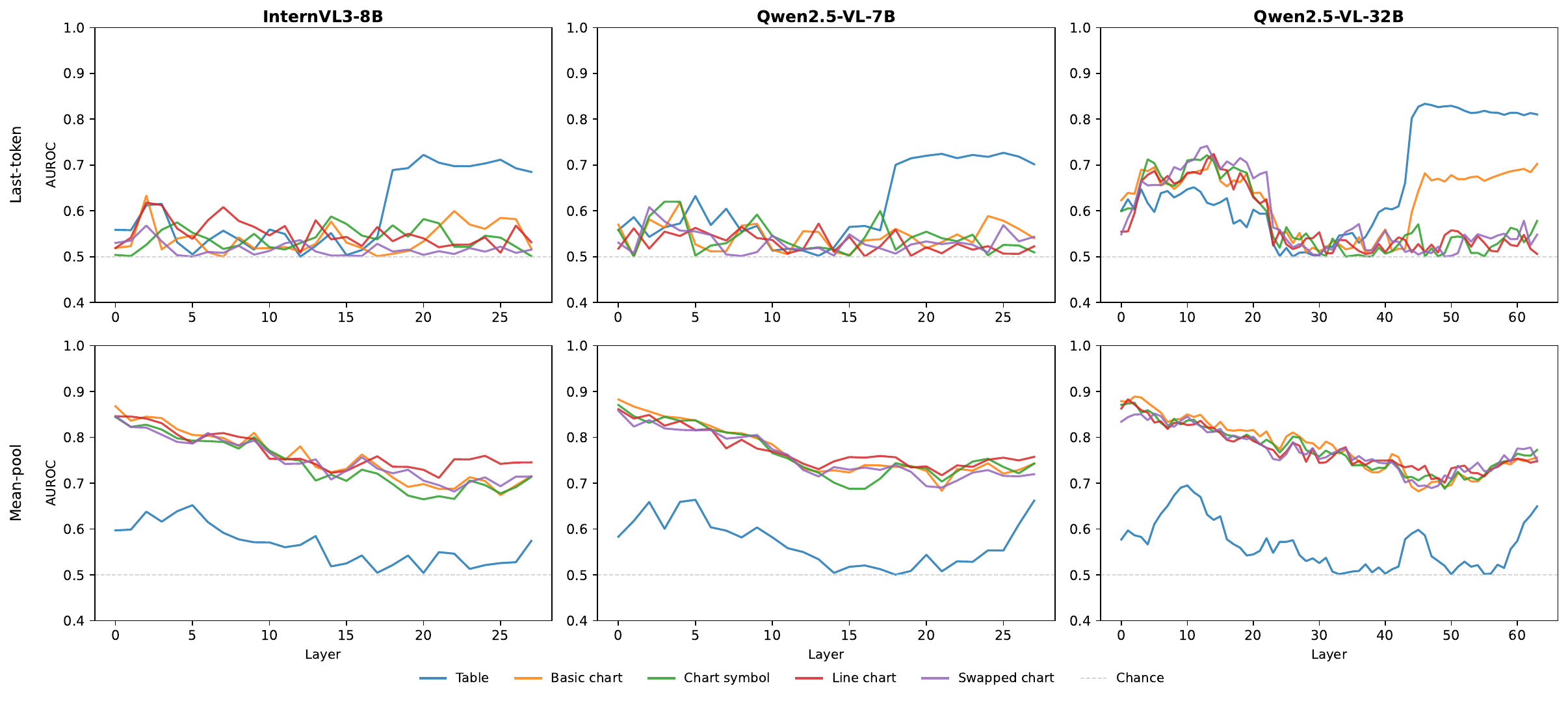}
\caption{Probe AUROC by layer for all three models on SciTabAlign+.
Top row: last-token probing. Bottom row: mean-pool probing. In all
models, table evidence (blue) rises sharply in later layers under
last-token probing while chart conditions remain near chance over the same range. Under
mean-pool probing, chart conditions are consistently decodable across
all layers, confirming that chart-relevant signal is present in the
model but does not reach the prediction position.}
\label{fig:probing-curves-all}
\end{figure*}

\begin{figure*}[t]
    \centering
    \includegraphics[width=\textwidth]{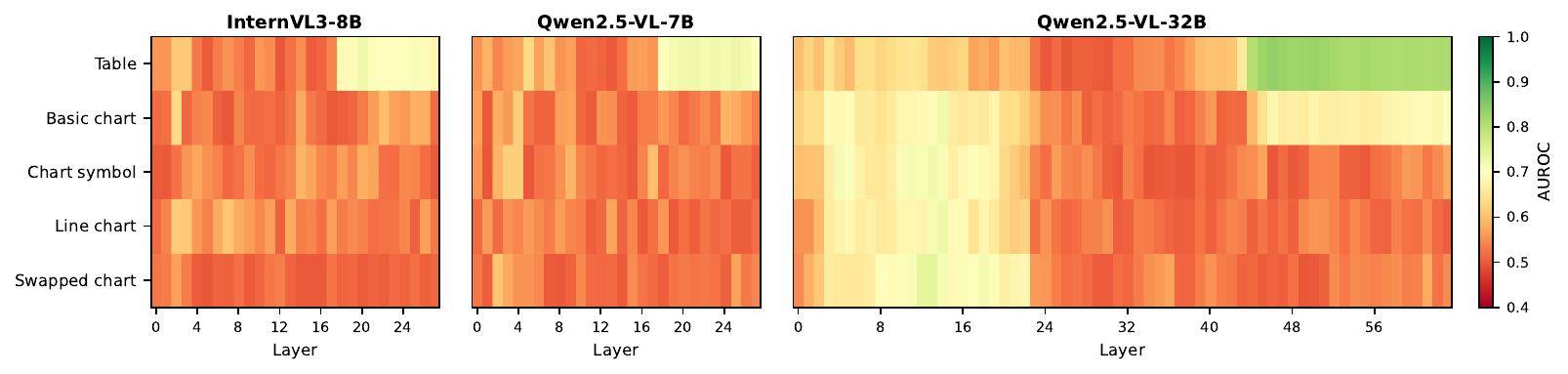}
    \caption{Last-token probe AUROC heatmap across layers and formats for all models. Table evidence (top row) accumulates to high AUROC in deeper layers and chart variants remain near chance across all three models.}
    \label{fig:probing-heatmap-all}
\end{figure*}

As shown in Figure~\ref{fig:probing-curves-all}, table evidence rises sharply in final layers under last-token probing while chart variants decline towards chance. Under mean-pool probing, chart-relevant signal remains decodable across all layers while tables perform worse than charts. Figure~\ref{fig:probing-heatmap-all} visualizes this pattern for last-token probe as a heatmap. 

\subsection{Image-Token Attention Distribution}
\label{app:internvl}

\begin{figure}[H]
    \centering
    \includegraphics[width=\columnwidth]{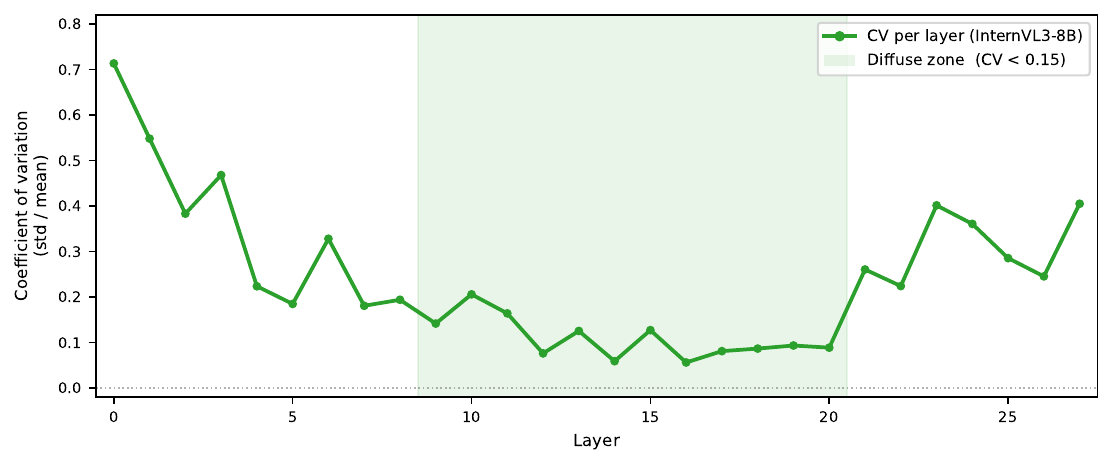}
    \caption{Coefficient of variation in image-token attention across  InternVL3-8B layers. Attention diffuses from concentrated (early) to distributed (deeper layers), ruling out summary-token bottlenecks.}
    \label{fig:internvl_cv}
\end{figure}

InternVL3-8B maintains near-proportional aggregate image-token attention (93\%), unlike Qwen models (4–11\%). We analyze whether this reflects distributed routing or concentration in summary tokens. Image-token attention in InternVL3-8B is distributed across positions rather than concentrated in summary tokens (Gini = 0.18, CV = 0.41 across 10 random claims). Figure~\ref{fig:internvl_cv} shows that within-image attention concentration drops from 0.71 in early layers to < 0.15 in deeper layers, confirming distributed visual routing.
Notably, this near-proportional image attention does not translate to better chart accuracy. InternVL3-8B achieved lower chart accuracy than both Qwen models, as shown in Table~\ref{tab:baseline-acc}, indicating that attending to chart evidence does not guarantee effective use of its content. Table~\ref{tab:dispersion} compares this distribution across models. The Qwen models show the opposite pattern. The bottom 90\% of image tokens receive only 25–34\% of image attention, concentrated near the start of the image token sequence.

\begin{table}[H]
\centering
\small
\begin{tabular}{lcc}
\toprule
\textbf{Model} & \textbf{Bottom-90\% share} & \textbf{Gini} \\
\midrule
Qwen2.5-VL-7B  & 25--34\% & 0.84 \\
Qwen2.5-VL-32B & 25--34\% & 0.79 \\
InternVL3-8B   & 80--82\% & 0.18 \\
\bottomrule
\end{tabular}
\caption{Distribution of attention within image tokens, over the same claims as Figure~\ref{fig:internvl_cv}.}
\label{tab:dispersion}
\end{table}

\section{Cross-format Representation Convergence}
\label{app:convergence}

\begin{figure*}[h]
    \centering
    \includegraphics[width=\textwidth]{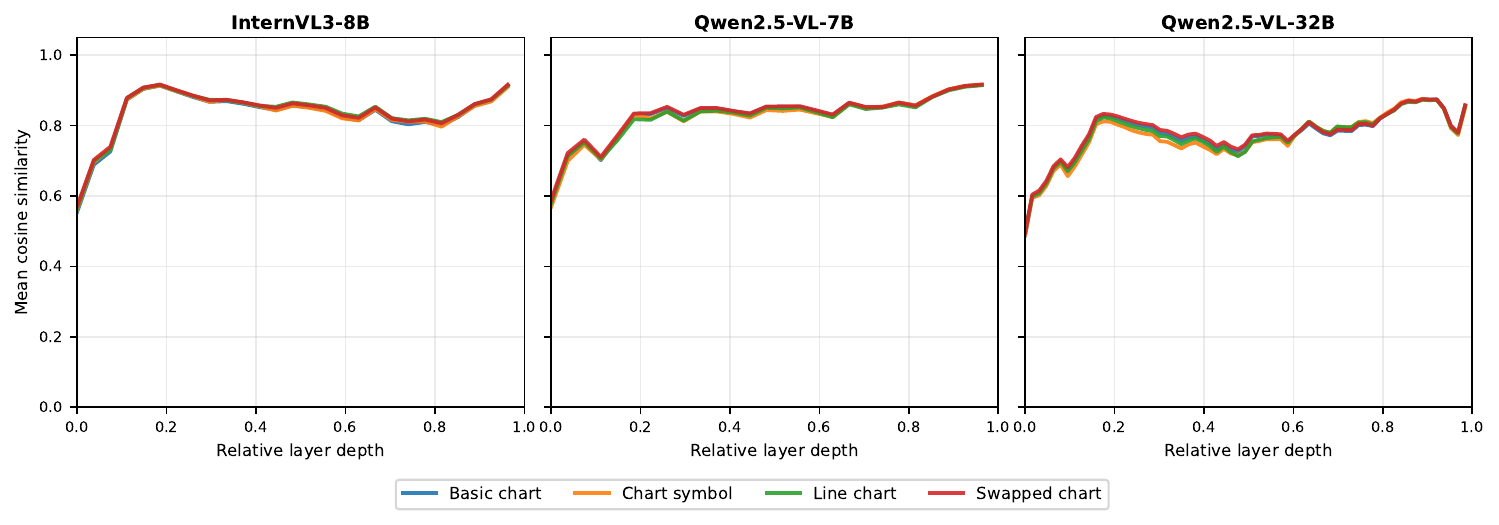}
    \caption{Mean cosine similarity between table and chart mean-pooled representations across all layers for each model (excluding final LM head layer). Table and chart representations start with low similarity in early layers but converge closer in deeper layers across all models and chart variants.}
    \label{fig:convergence}
\end{figure*}

As shown in Figure~\ref{fig:convergence}, table and chart representations converge to high cosine similarity in deeper layers across all models.

\section{Chain-of-Thought Ablation}
\label{app:cot}

The prompt template is shown in Figure~\ref{app:cot-prompt}. As shown in Table~\ref{tab:cot-comparison}, chain-of-thought prompting degrades performance for Qwen models while improving InternVL3-8B.

\begin{table}[H]
    \centering
    \resizebox{\columnwidth}{!}{%
    \begin{tabular}{lrrrrrr}
    \toprule
    & \multicolumn{2}{c}{\textbf{Standard}}
    & \multicolumn{2}{c}{\textbf{CoT}}
    & \multicolumn{2}{c}{\textbf{$\Delta$}} \\
    \cmidrule(lr){2-3}\cmidrule(lr){4-5}\cmidrule(lr){6-7}
    \textbf{Model} & Acc & F1 & Acc & F1 & $\Delta$Acc & $\Delta$F1 \\
    \midrule
    Qwen2.5-VL-7B  & 66.0 & 65.9 & 45.7 & 52.4 & $-20.3$ & $-13.5$ \\
    Qwen2.5-VL-32B & 68.5 & 68.2 & 63.6 & 64.9 & $-4.9$  & $-3.3$  \\
    InternVL3-8B   & 56.2 & 55.2 & 59.9 & 59.8 & $+3.7$  & $+4.6$  \\
    \bottomrule
    \end{tabular}%
    }
    \caption{Effect of chain-of-thought prompting on basic chart accuracy and macro-F1 (\%) on SciTabAlign+. $\Delta$ values are CoT minus standard.}
\label{tab:cot-comparison}
\end{table}

\begin{figure}
\noindent\fbox{%
\parbox{0.97\columnwidth}{%
\centering\textbf{Chain-of-Thought Prompt Template}\\[0.5em]
\small
\begin{flushleft}
\textbf{System:} You are an expert in claim verification against scientific papers.\\[0.5em]
 
\textbf{User:} [image]\\
\textbf{Step 1:} Carefully examine the chart. Describe the data you observe --- list specific values, comparisons between groups, and any trends visible. Be precise and mention exact numbers shown on axes or labels if you can read them.\\[0.5em]
 
\textbf{Step 2:} Based ONLY on what you described in Step 1, determine whether the claim is Supported or Refuted.\\[0.5em]
 
\textbf{Claim:} \texttt{\{claim\}}\\[0.3em]
 
Think step by step. Format your final answer as: \texttt{<ans> YOUR ANSWER </ans>}
\end{flushleft}
}%
}
\caption{Chain-of-thought prompt template for claim verification against chart format.}
\label{app:cot-prompt}
\end{figure}

\section{Claim-Only Baseline}
\label{app:claim-only}

Table~\ref{tab:claim-only} shows claim-only probes reach 58--67\% AUROC vs. 84--89\% for charts. This gap shows that the mean-pool probe captures visual information, and not text leakage.

\begin{table}[ht]
    \centering
    \resizebox{\columnwidth}{!}{%
    \begin{tabular}{lrrr}
    \toprule
    \textbf{Model} & \textbf{Claim-Only} & \textbf{Basic Chart} & \textbf{$\Delta$} \\
    \midrule
    Qwen2.5-VL-7B & 58.5\% & 88.3\% & +29.8\% \\
    Qwen2.5-VL-32B & 67.4\% & 88.9\% & +21.5\% \\
    InternVL3-8B & 62.0\% & 86.8\% & +24.8\% \\
    \bottomrule
    \end{tabular}%
    }
    \caption{Claim-only vs. basic chart mean-pool AUROC. Claim-only probe is trained only on claim text (no image tokens provided). The large gap ($\Delta$) confirms that probe performance is driven by visual information, not just claim-text cues.}
    \label{tab:claim-only}
\end{table}

\section{Error Analysis}
\label{app:error-analysis}

\begin{figure*}[t]
\centering

\begin{tcolorbox}[
  enhanced,
  colback=claimgold,
  colframe=claimframe,
  boxrule=0.7pt,
  arc=3pt,
  left=8pt,
  right=8pt,
  top=5pt,
  bottom=5pt,
  fontupper=\small,
  width=\textwidth
]
\textbf{Claim:}
``Among all the baselines, GDPL does not obtain the most preference against PPO.''
\hfill
\textbf{True Label:}~\textit{Refuted}
\end{tcolorbox}

\vspace{0.5em}

\begin{minipage}[t]{0.49\textwidth}
\begin{tcolorbox}[
  enhanced,
  title={\small\bfseries Chart Evidence\hfill
  \textcolor{incorrect}{\texttimes~Model}~~%
  \textcolor{correct}{\checkmark~Probe}},
  colback=chartbg,
  colframe=chartframe,
  colbacktitle=chartframe!25,
  coltitle=black,
  boxrule=0.8pt,
  arc=3pt,
  left=6pt,
  right=6pt,
  top=5pt,
  bottom=6pt
]

\begin{center}
\includegraphics[width=0.82\linewidth]{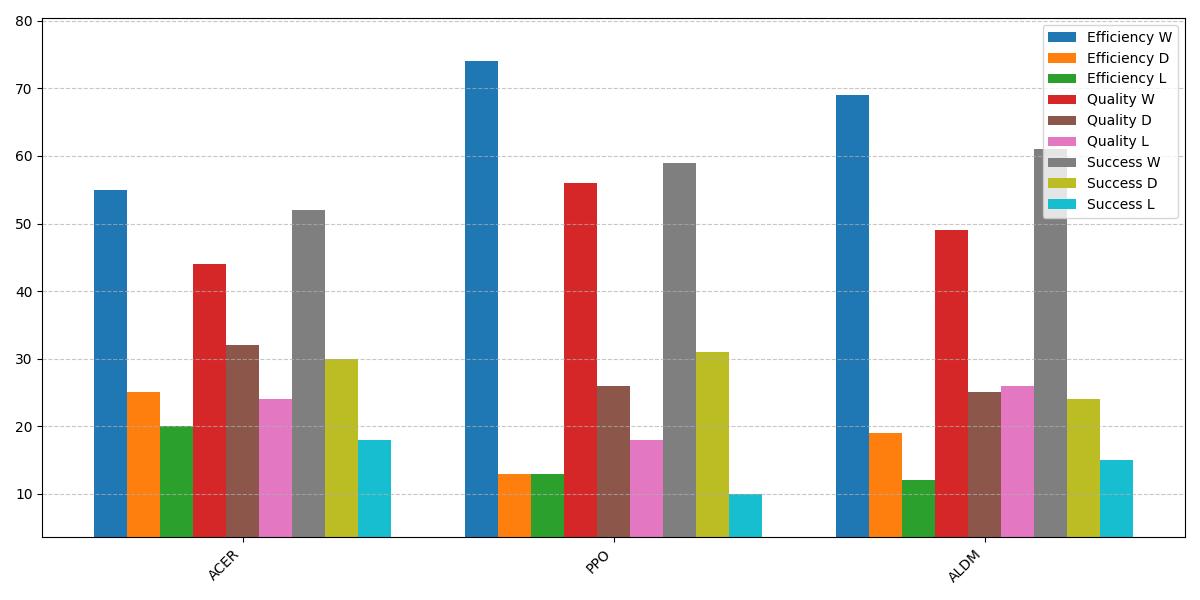}
\end{center}

{\scriptsize\textit{Caption: The count of human preference on dialog session pairs that GDPL wins (W), draws with (D) or loses to (L) other methods based on different criteria. One method wins the other if the majority prefer the former one.}}

\vspace{0.6em}

\begin{tabular}{@{}p{0.22\linewidth}l@{}}
\textbf{Model:} & \incorrectbadge{Supported} \\[3pt]
\textbf{Probe:} & \correctbadge{Refuted} \\
\end{tabular}

\vspace{0.6em}

\textbf{\small Model response (excerpt):}
\begin{quote}\footnotesize
``GDPL and PPO have the same highest value in the `Success W'
category, neither is definitively preferred over the other
[\ldots] the claim is supported.''
\end{quote}

\small
The model hallucinates GDPL's presence in the chart despite it
not appearing as an explicit bar. The mean-pool probe correctly decodes
the \textit{Refuted} label.

\end{tcolorbox}
\end{minipage}
\hfill
\begin{minipage}[t]{0.49\textwidth}
\begin{tcolorbox}[
  enhanced,
  title={\small\bfseries Table Evidence\hfill
  \textcolor{correct}{\checkmark~Model}~~%
  \textcolor{incorrect}{\texttimes~Probe}},
  colback=tablebg,
  colframe=tableframe,
  colbacktitle=tableframe!20,
  coltitle=black,
  boxrule=0.8pt,
  arc=3pt,
  left=6pt,
  right=6pt,
  top=5pt,
  bottom=6pt
]

\begin{center}
\resizebox{0.96\linewidth}{!}{%
\begin{tabular}{lrrrrrrrrr}
\toprule
\textbf{VS.}
& \multicolumn{3}{c}{\textbf{Efficiency}}
& \multicolumn{3}{c}{\textbf{Quality}}
& \multicolumn{3}{c}{\textbf{Success}} \\
\cmidrule(lr){2-4}
\cmidrule(lr){5-7}
\cmidrule(lr){8-10}
& W & D & L & W & D & L & W & D & L \\
\midrule
ACER & 55 & 25 & 20 & 44 & 32 & 24 & 52 & 30 & 18 \\
PPO  & 74 & 13 & 13 & 56 & 26 & 18 & 59 & 31 & 10 \\
ALDM & 69 & 19 & 12 & 49 & 25 & 26 & 61 & 24 & 15 \\
\bottomrule
\end{tabular}}
\end{center}

{\scriptsize\textit{Caption: The count of human preference on dialog session pairs that GDPL wins (W), draws with (D) or loses to (L) other methods based on different criteria. One method wins the other if the majority prefer the former one.}}

\vspace{0.6em}

\begin{tabular}{@{}p{0.22\linewidth}l@{}}
\textbf{Model:} & \correctbadge{Refuted} \\[3pt]
\textbf{Probe:} & \incorrectbadge{Supported} \\
\end{tabular}

\vspace{0.6em}

\textbf{\small Model response (excerpt):}
\begin{quote}\footnotesize
``GDPL wins 55 times and loses 13 times in efficiency [\ldots]
GDPL wins more often than it loses in all three criteria,
the claim is refuted.''
\end{quote}

\small
The model correctly reads the structured values and resolves double negation in the claim, and makes correct prediction. The probe predicts \textit{Supported} incorrectly, consistent with the pattern observed across negative claims in the samples instances for this analysis.

\end{tcolorbox}
\end{minipage}

\caption{Representative error analysis example for Qwen2.5-VL-7B.
\textbf{Left:} Chart evidence where the model fails through hallucination
while the probe succeeds.
\textbf{Right:} Table evidence where the model succeeds through negation
handling while the probe fails.}
\label{fig:error-example}
\end{figure*}

We randomly selected 5 instances each from cases where the model fails but probe succeeds on basic chart evidence, and where models succeeds but probe fails on table evidence for Qwen2.5-VL-7B, and examined the model responses qualitatively. Chart failures are predominantly driven by hallucinated content, where the model fabricates values or entities not present in the chart. Cases where the model outperforms the probe on table evidence are dominated by negative claims, where the model correctly resolves negation while the probe predicts incorrectly. Figure~\ref{fig:error-example} shows a representative example.

\end{document}